\documentclass{article}
\usepackage{spconf,amsmath,graphicx,hyperref}
\usepackage{booktabs}
\usepackage{amssymb}
\usepackage{makecell}
\usepackage{multirow}
\usepackage{tabularx}
\usepackage{array}
\usepackage{enumitem}

\usepackage{multirow}
\usepackage{booktabs}
\usepackage{makecell}
\usepackage[table,svgnames]{xcolor}

\title{Depth-Guided Metric-Aware Temporal Consistency for Monocular Video Human Mesh Recovery}
\name{Jiaxin Cen$^{1}$, Xudong Mao$^{1}$, Guanghui Yue$^{2}$, Wei Zhou$^{3}$, Ruomei Wang$^{1}$, Fan Zhou$^{1}$, Baoquan Zhao$^{1,*}$\thanks{$^{*}$ To whom correspondence should be addressed.
}
\thanks{This work was supported in part by Shenzhen Medical
Research Fund under Grant A2403035, in part by the Project of Department
of Education of Guangdong Province under Grant 2025KTSCX111, in part by
the National Natural Science Foundation of China under Grant 62371305 and 62302535, in part by Guangdong Basic and Applied Basic Research Foundation under Grant 2024A1515030025 and 2023A1515011639.}}

\address{$^{1}$Sun Yat-sen University, China \hspace{10pt}
$^{2}$Shenzhen University, China \hspace{10pt} $^{3}$Cardiff University, UK}

\begin{document}
\maketitle
\begin{abstract}
Monocular video human mesh recovery faces fundamental challenges in maintaining metric consistency and temporal stability due to inherent depth ambiguities and scale uncertainties. While existing methods rely primarily on RGB features and temporal smoothing, they struggle with depth ordering, scale drift, and occlusion-induced instabilities. We propose a comprehensive depth-guided framework that achieves metric-aware temporal consistency through three synergistic components: A \emph{Depth-Guided Multi-Scale Fusion} module that adaptively integrates geometric priors with RGB features via confidence-aware gating; A \emph{Depth-guided Metric-Aware Pose and Shape} (D-MAPS) estimator that leverages depth-calibrated bone statistics for scale-consistent initialization; A \emph{Motion-Depth Aligned Refinement} (MoDAR) module that enforces temporal coherence through cross-modal attention between motion dynamics and geometric cues. Our method achieves superior results on three challenging benchmarks, demonstrating significant improvements in robustness against heavy occlusion and spatial accuracy while maintaining computational efficiency.
\end{abstract}


\begin{figure*}[!t]
\includegraphics[page=1, width=1\linewidth]{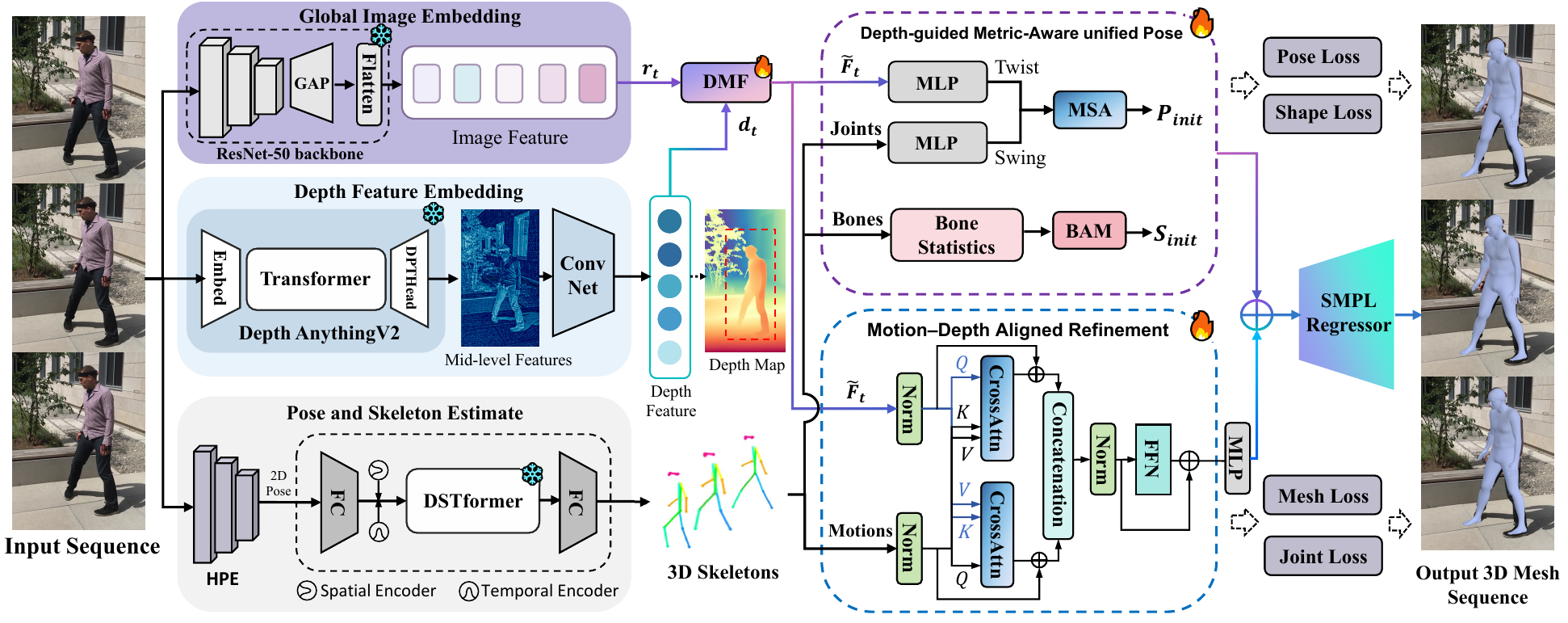}
\caption{
\textbf{Overall architecture of our framework.}
}
\label{fig:pipeline}
\end{figure*}

\section{Introduction}
\label{sec:intro}

Recovering temporally consistent 3D human meshes from monocular video sequences remains one of the most challenging problems in computer vision, with critical applications spanning virtual reality, motion capture, and human behavior analysis. Despite significant advances in single-frame human mesh reconstruction~\cite{bogo2016keep,kanazawa2018end,kolotouros2019learning,kocabas2021pare}, extending these methods to video sequences introduces fundamental challenges that purely RGB-based approaches struggle to address effectively.
The core difficulty lies in the inherent ill-posedness of monocular 3D reconstruction: infinite 3D configurations can project to identical 2D observations, creating persistent ambiguities in depth ordering, metric scale, and temporal consistency. 

Existing video-based methods such as VIBE, TCMR, and GLoT~\cite{kocabas2020vibe,choi2021beyond,shen2023global} attempt to resolve these ambiguities through temporal modeling and motion priors. However, they remain fundamentally limited by their reliance on appearance features alone, leading to characteristic failure modes including scale drift, depth ordering errors, and temporal jitter under challenging conditions such as rapid motion, occlusion, and viewpoint changes.
At a fundamental level, the ill-posed nature of monocular 3D reconstruction means that multiple 3D configurations can project to identical 2D observations when depth information along camera rays is ambiguous. Consequently, temporal smoothing alone cannot resolve scale and depth ordering ambiguities~\cite{ishii20243d}. Recent approaches have explored various strategies to improve temporal stability. Semi-analytical methods like ARTS~\cite{tang2024arts} decompose the problem into interpretable components, while co-evolution frameworks like PMCE~\cite{you2023co} model pose-shape interactions explicitly. However, these methods still operate primarily in the RGB domain and inherit fundamental depth-related limitations~\cite{li2021hybrik}.

The emergence of robust monocular depth estimation models presents a compelling opportunity to address these limitations through explicit geometric reasoning. Monocular depth estimation offers several advantages for human mesh reconstruction: \emph{Metric awareness} of person-camera distance and bone length regularization to mitigate scale drift; \emph{Depth ordering enforcement} via along-ray priors, reducing front-back ambiguity and improving occlusion reasoning; \emph{Geometric delineation} through depth discontinuities for foreground-background separation~\cite{chen2016single}. With the emergence of robust depth foundation models (e.g., Depth Anything~\cite{yang2024depth}) that generalize across diverse scenes and camera configurations, incorporating depth information becomes increasingly viable for improving 3D accuracy and 2D alignment~\cite{wang2025blade,ranftl2021vision}.
However, naively incorporating depth estimates introduces new challenges. Raw depth predictions can be noisy, miscalibrated, or inconsistent across frames, potentially degrading rather than improving reconstruction quality. Furthermore, effectively fusing depth information with existing RGB-based pipelines requires careful design to leverage geometric constraints while maintaining robustness to depth estimation errors.
Our comprehensive evaluation demonstrates that this depth-guided approach achieves state-of-the-art performance across multiple benchmarks while addressing the fundamental limitations of purely RGB-based methods. 
In summary, we make the following contributions:
\begin{itemize}[leftmargin=*,topsep=2pt,itemsep=2pt,parsep=0pt,partopsep=0pt]
  \item We propose a depth-guided multi-scale fusion architecture that adaptively integrates geometric priors with RGB features through confidence-aware gating, enabling robust performance under depth estimation uncertainties.
  \item We introduce D-MAPS, a metric-aware initialization strategy that leverages depth-calibrated bone length statistics to achieve scale-consistent pose and shape estimation across temporal sequences.
  \item We develop MoDAR, a cross-modal refinement framework that aligns motion dynamics with geometric constraints through bidirectional attention, ensuring temporal coherence without over-smoothing.
\end{itemize}

\section{Method}
\label{sec:method}

\subsection{RGB-Depth Feature Extraction and Fusion}

Given input video $\{I_t\}_{t=1}^{T}$, we extract RGB features $r_t$ using ResNet-50~\cite{he2016deep} and depth features using Depth Anything v2 (DAv2)~\cite{yang2024depth} as a pretrained feature extractor. Following BLADE~\cite{wang2025blade}, we avoid directly using depth values due to noise and calibration issues. Instead, we extract intermediate activations from the DAv2 encoder $E_{\mathrm{DA}}$ at stage $k^*$ and refine them through lightweight convolutions (denoted by $\psi$), followed by upsampling $U(\cdot)$:
\begin{equation}
d_t = U\!\big(\psi\!\big(E_{\mathrm{DA}}^{(k^{\ast})}(I_t)\big)\big)
\end{equation}

\begin{figure}[!t]
\centering
\includegraphics[width=1\linewidth]{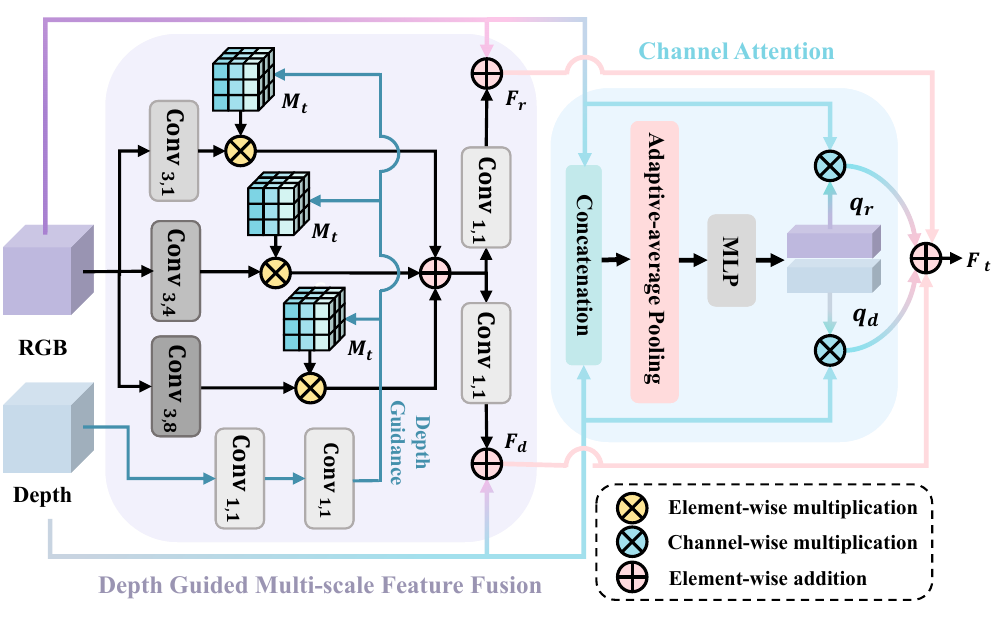}
\caption{
\textbf{Depth-Guided Multi-Scale Fusion module.}
}
\label{fig:fusion_module}
\end{figure}

For robust multi-modal fusion \cite{hu2018squeeze}, the depth pathway generates a modulation mask $M_t \in [0,1]^{H\times W\times C}$ via two $1\times1$ layers and sigmoid activation. This mask modulates RGB features element-wise to form the RGB stream $F_r$, while the processed depth features form the depth stream $F_d$. Channel-wise calibration through adaptive pooling and an MLP produces gates $\mathbf{q}_r, \mathbf{q}_d \in \mathbb{R}^{C}$ for dynamic modality balancing. The detailed structure of this fusion module is illustrated in Figure \ref{fig:fusion_module}. The fused feature $\tilde{F}_t$ is obtained by concatenation and a projection head $\phi(\cdot)$:
\begin{equation}
\tilde{F}_t = \phi\!\big([\mathbf{q}_r \odot F_r \| \mathbf{q}_d \odot F_d]\big)
\end{equation}

\subsection{D-MAPS: Unified Metric-Aware Pose-Shape Initialization}

D-MAPS produces a metric-consistent initialization of SMPL parameters, including pose $p_{\text{init}}$ and shape $s_{\text{init}}$, by coupling kinematics with depth-calibrated bone statistics under reliability gating and temporal smoothing. Joint rotations are decomposed into swing components derived from normalized bone directions and twist components predicted from $\tilde{F}_t$ with local depth patches. These components are temporally aggregated through lightweight self-attention to yield $p_{\text{init}}$.

Concurrently, metric bone lengths are estimated as depth-confidence-weighted temporal averages and fused with template statistics through a learned depth gate to obtain calibrated bone lengths $B^{Z}$. The rest-pose template is then scaled along the kinematic tree, and an analytically initialized MLP regresses $s_{\text{init}}$.

We define the temporal weight as $w_t = \sigma(\eta \overline{m}_t)$, where $\overline{m}_t$ represents the mean depth confidence within the person region of interest. The fusion gate is defined as $\alpha = \sigma\big(\eta \frac{1}{T}\sum_{t=1}^{T}\overline{m}_t\big)$. The temporal bone length estimation is:
\begin{equation}
\tilde{B}_{(i,j)} = \frac{\sum_{t=1}^{T} w_t \|S^{3D}_{t,i} - S^{3D}_{t,j}\|}{\sum_{t=1}^{T} w_t}
\end{equation}
\begin{equation}
B^{Z}_{(i,j)} = \alpha \tilde{B}_{(i,j)} + (1-\alpha) \bar{B}_{(i,j)}
\end{equation}

Additionally, detected 2D joints $K_t \in \mathbb{R}^{J\times2}$ are root-centered and scale-normalized, then lifted to pelvis-centered 3D joints $\hat{J}_t \in \mathbb{R}^{J\times3}$ using MotionBERT's DSTformer~\cite{zhu2023motionbert}. These skeletal representations regularize D-MAPS during initialization and serve as motion tokens for subsequent MoDAR refinement.

\subsection{MoDAR: Motion-Depth Aligned Refinement}

While initialization provides metric scale and depth ordering, temporal inconsistencies and occlusion artifacts may remain. MoDAR refines parameters by aligning motion with depth-enhanced representations, transforming $(p_{\text{init}}, s_{\text{init}})$ into $(p_{\text{ref}}, s_{\text{ref}})$.

Motion tokens serve as queries, while fused tokens $\tilde{F}_t$ act as keys and values in a cross-modal attention mechanism. Two stacked cross-attention blocks enable bidirectional information flow between motion and fusion streams. Layer normalization and a compact feed-forward network produce a context feature $F'$ that emphasizes geometry-supported motion while suppressing static background clutter.

Lightweight residual heads update the parameters, and a causal temporal filter applied to residuals mitigates high-frequency oscillations:
\begin{equation}
\begin{aligned}
\mathbf{x}_t &= (1-\rho)\mathbf{x}_{t-1} + \rho\left(\mathbf{x}_0 + \mathbf{g}_t \odot \Delta\mathbf{x}(F'_t)\right) \\
\mathbf{x}_t &= \begin{bmatrix}p_t\\ s_t\end{bmatrix}, \quad \mathbf{g}_t = \sigma(W F'_t)
\end{aligned}
\label{eq:refine_ps_compact}
\end{equation}

\subsection{Training Objectives}

Our training strategy employs standard HMR supervision with multi-level objectives. We optimize mesh reconstruction, joint positioning, pose parameters, and shape parameters, complemented by a temporal smoothness term to reduce jitter. Depth information acts solely as a feature cue without requiring depth-specific loss functions. The total loss function is:
\begin{align}
\mathcal{L} &= \lambda_m \mathcal{L}_{\mathrm{mesh}} + \lambda_j \mathcal{L}_{\mathrm{joint}} + \lambda_p \mathcal{L}_{\mathrm{pose}} \nonumber\\
&\quad + \lambda_s \mathcal{L}_{\mathrm{shape}} + \lambda_t \mathcal{L}_{\mathrm{smooth}}
\label{eq:total_loss_modar}
\end{align}

We employ a two-phase training schedule. Phase 1 focuses on warming up the backbone and RGB-depth fusion under motion supervision with gated depth integration for stability. Phase 2 trains D-MAPS, MoDAR, and the SMPL regressor end-to-end with multi-step learning rate decay and delayed regularization terms.

\begin{figure}[t]
\centering
\includegraphics[width=\columnwidth]{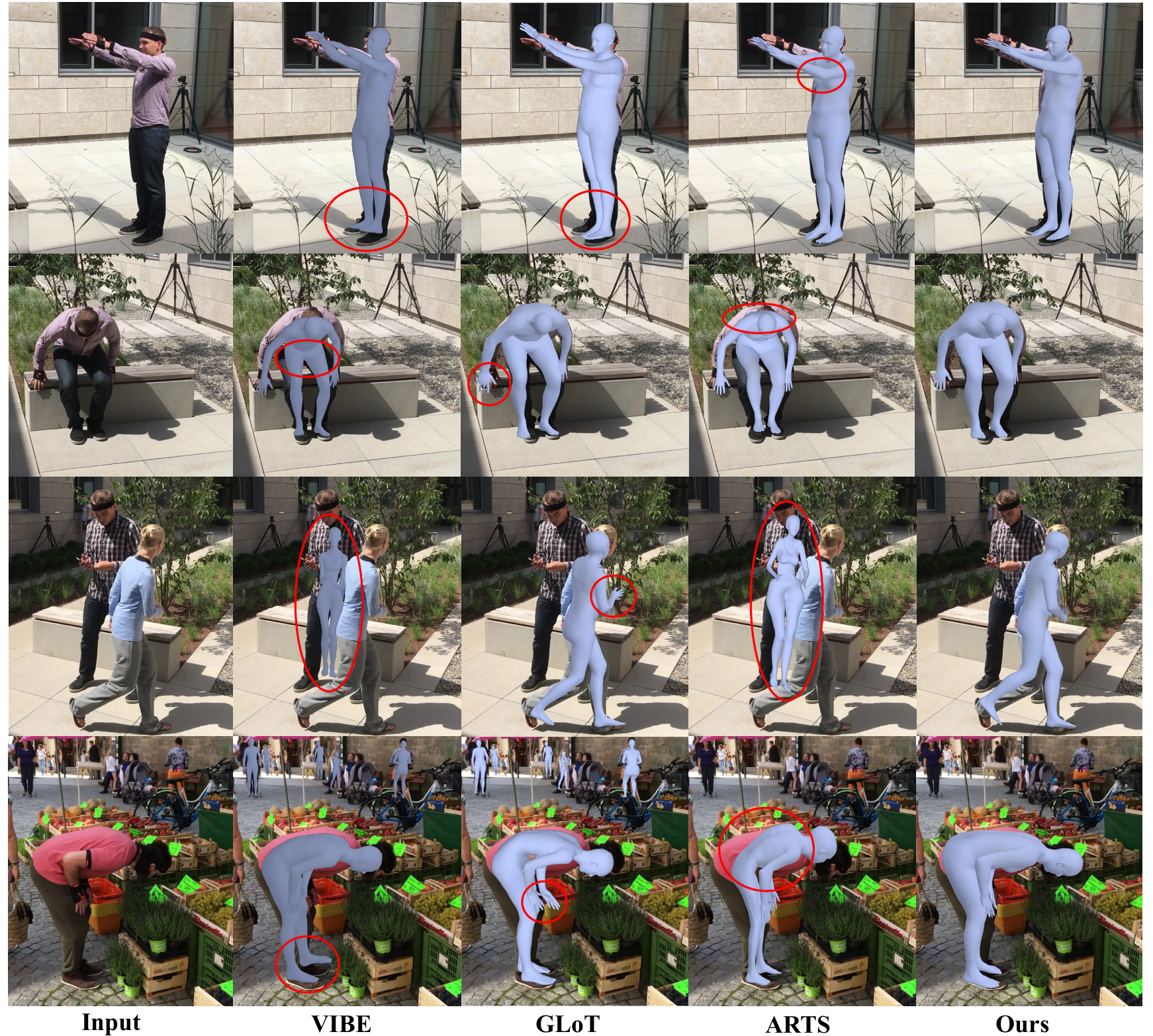}
\caption{\textbf{Qualitative comparison on challenging in-the-wild videos.}
From left to right: Input frame, VIBE~\cite{kocabas2020vibe}, GLoT~\cite{shen2023global}, ARTS~\cite{tang2024arts}, and our method.
Our depth-guided approach demonstrates superior metric consistency, reduced depth ordering errors, and improved limb articulation in complex scenarios including crouching and interaction scenes.}
\label{fig:qualitative}
\end{figure}

\begin{table*}[ht]
\centering
\caption{Quantitative comparison on standard benchmarks. Metrics include MPJPE and PA-MPJPE (mm), MPVPE (mm; 3DPW only), and Accel (mm/s²). Best results in \textbf{bold}; second best \underline{underlined}.}
\renewcommand{\arraystretch}{1.25}
\setlength{\tabcolsep}{5pt}
\resizebox{\textwidth}{!}{
\begin{tabular}{l|cccc|ccc|ccc}
\Xhline{1.2pt}
\multirow{2}{*}{Method} 
& \multicolumn{4}{c|}{\textbf{3DPW}} 
& \multicolumn{3}{c|}{\textbf{Human3.6M}} 
& \multicolumn{3}{c}{\textbf{MPI-INF-3DHP}} \\
\cline{2-11}
& MPJPE $\downarrow$ & PA-MPJPE $\downarrow$ & MPVPE $\downarrow$ & Accel $\downarrow$
& MPJPE $\downarrow$ & PA-MPJPE $\downarrow$ & Accel $\downarrow$
& MPJPE $\downarrow$ & PA-MPJPE $\downarrow$ & Accel $\downarrow$ \\
\Xhline{0.8pt}
VIBE (CVPR'20) \cite{kocabas2020vibe}
& 95.12 & 58.44 & 101.26 & 24.70
& 77.24 & 54.06 & 26.72
& 105.08 & 69.84 & 28.06 \\
MEAD (ICCV'21) \cite{wan2021encoder}
& 82.37 & \underline{46.88} & 95.12 & 18.64
& 57.64 & 39.08 & 5.86
& 84.66 & 57.14 & 8.52 \\
GLoT (CVPR'23) \cite{shen2023global}
& 81.42 & 51.63 & 97.51 & 7.32
& 66.05 & 45.74 & 4.68
& 95.76 & 62.02 & 8.08 \\
PMCE (ICCV'23) \cite{you2023co}
& 74.76 & 47.05 & 86.31 & \underline{6.96}
& 54.58 & 37.26 & \textbf{3.02}
& 80.46 & 55.06 & 8.00 \\
ARTS (MM'24) ~\cite{tang2024arts}
& \underline{72.15} & 47.96 & \underline{85.47} & \textbf{6.91}
& \underline{51.55} & \underline{36.63} & 4.11
& \underline{75.45} & \underline{54.21} & \underline{$7.90$} \\
\rowcolor{Lavender!55}
\textbf{DMAPS (Ours)}
& \textbf{69.31} & \textbf{46.68} & \textbf{82.61} & 7.14
& \textbf{51.18} & \textbf{35.96} & \underline{3.92}
& \textbf{73.45} & \textbf{53.87} & \textbf{7.89} \\
\Xhline{1.2pt}
\end{tabular}}
\label{tab:sota_benchmark}
\end{table*}

\begin{table}[!t]
\renewcommand{\arraystretch}{1.25}
\caption{Ablation study on 3DPW test set. Components are added cumulatively, while single-module controls isolate the contribution of D-MAPS and MoDAR. Best results in \textbf{bold}; second best \underline{underlined}.}
\centering
\begingroup
\footnotesize
\setlength{\tabcolsep}{3pt}
\begin{tabularx}{\columnwidth}{@{}l *{4}{>{\centering\arraybackslash}X}@{}}
\Xhline{1.2pt}
Configuration & {\scriptsize MPJPE} & {\scriptsize PA-MPJPE} & {\scriptsize MPVPE} & {\scriptsize Accel} \\
& {\scriptsize ($\downarrow$)} & {\scriptsize ($\downarrow$)} & {\scriptsize ($\downarrow$)} & {\scriptsize ($\downarrow$)} \\
\Xhline{0.8pt}
RGB-only baseline              & 82.45 & 58.40 & 85.50 & 9.20 \\
+ Mask-guided fusion           & 73.12 & 53.80 & \underline{82.34} & 8.60 \\
+ Quality-aware depth          & \underline{71.05} & \textbf{46.10} & 82.70 & \underline{7.12} \\
+ D-MAPS (w/o MoDAR)           & 72.20 & 47.10 & 86.50 & 7.28 \\
+ MoDAR (w/o D-MAPS)           & 71.36 & 48.48 & 84.24 & 7.32 \\
\rowcolor{Lavender!55}
\textbf{Ours (complete)}       & \textbf{69.48} & \underline{46.52} & \textbf{82.19} & \textbf{6.97} \\
\Xhline{1.2pt}
\end{tabularx}
\endgroup
\label{tab:ablation_monocular_single}
\end{table}

\section{Experiments}
\label{sec:exp}
\subsection{Experimental Setup}
We adopt standard mixed 2D–3D training: COCO and MPII for 2D supervision, and Human3.6M, MPI-INF-3DHP, 3DPW for 3D.
Evaluation is on the 3DPW test split, Human3.6M Protocol~2, and MPI-INF-3DHP protocol; we report MPJPE and PA-MPJPE (mm), MPVPE on 3DPW, and acceleration error (mm\,s$^{-2}$), using identical 2D detections and the same lifting~\cite{zhu2023motionbert} with ResNet-50 features~\cite{he2016deep}.

\begin{figure}[!t]
\centering
\includegraphics[width=1\linewidth]{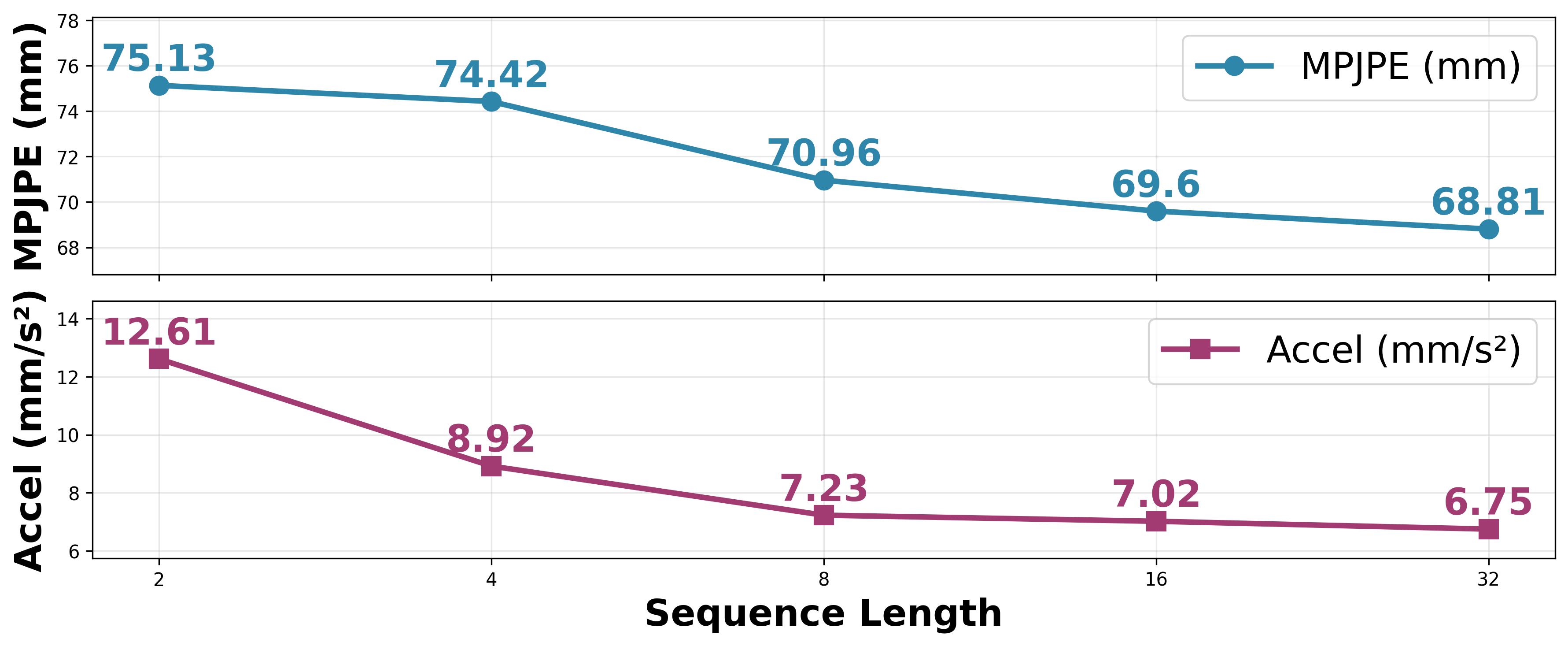}
\caption{\textbf{Impact of sequence length on D-MAPS performance.} Longer sequences supply richer temporal context, consistently lowering MPJPE and Accel.}
\label{fig:d_maps_seq_trends}
\end{figure}

\subsection{Comparison with State-of-the-Art Methods}
Table~\ref{tab:sota_benchmark} presents comprehensive quantitative evaluation results across three standard public benchmarks. Our method achieves superior spatial accuracy across all datasets, particularly on the challenging 3DPW where we obtain significant margins in both joint and vertex errors. While certain methods exhibit marginally lower acceleration errors, this is often attributed to excessive smoothing that compromises rapid motion recovery. In contrast, our approach maintains competitive temporal stability while effectively preserving high-frequency motion details and geometric fidelity.

The qualitative study in Figure~\ref{fig:qualitative} shows consistent gains across extended-arm, crouching, interaction, and bending scenes, preserving contacts, enforcing scale consistency and along-ray ordering under occlusion, and tracking large articulations without oversmoothing to reduce floating, self-intersection, and limb flips. 

\subsection{Ablation Study and Analysis}
Table~\ref{tab:ablation_monocular_single} presents the ablation analysis. The integration of mask-guided fusion and quality-aware depth yields substantial performance gains, underscoring the necessity of robust foreground and depth cues for stable reconstruction.
We further isolate the contributions of D-MAPS and MoDAR using single-module controls. D-MAPS enforces metric consistency via bone-length priors, while MoDAR targets temporal coherence through motion-depth alignment. Notably, deploying either module in isolation results in slight performance regressions, as individual constraints can introduce conflicts without mutual support. In contrast, \textbf{Ours (complete)} leverages their synergy. Although rigid bone constraints cause a marginal increase in PA-MPJPE by limiting Procrustes alignment freedom, the complete model achieves the best performance in MPJPE, MPVPE, and Accel. This confirms that the improvements in absolute metric fidelity and temporal stability outweigh the minor trade-off in normalized alignment.
In addition, Fig.~\ref{fig:d_maps_seq_trends} shows that longer sequences monotonically reduce MPJPE and Accel, as richer temporal context improves pose accuracy and motion smoothness.

\section{Conclusion}

We present a comprehensive depth-guided framework for monocular video human mesh reconstruction, addressing fundamental challenges in scale ambiguity and temporal consistency. Our approach integrates three key innovations: depth-guided multi-scale fusion, metric-aware initialization (D-MAPS), and motion-depth aligned refinement (MoDAR). Extensive experiments validate that this framework effectively resolves ambiguities in extreme scenarios like severe occlusion, providing a principled solution for robust, temporally stable 3D human understanding.


\bibliographystyle{IEEEbib}
\bibliography{strings,refs}
\end{document}